\documentclass[conference]{IEEEtran}
\IEEEoverridecommandlockouts
\usepackage{cite}
\usepackage{amsmath,amssymb,amsfonts}
\usepackage{algorithmic}
\usepackage{graphicx}%
\graphicspath{{./img_folder/}}
\usepackage{textcomp}
\usepackage[export]{adjustbox}
\usepackage{xcolor}
\def\BibTeX{{\rm B\kern-.05em{\sc i\kern-.025em b}\kern-.08em
    T\kern-.1667em\lower.7ex\hbox{E}\kern-.125emX}}
\begin{document}

\title{Cross Spatial Temporal Fusion Attention for Remote Sensing Object Detection via Image Feature Matching\\
}

\author{
\IEEEauthorblockN{1\textsuperscript{st} Abu Sadat Mohammad Salehin Amit}
\IEEEauthorblockA{\textit{College of Computer Science and Technology} \\
\textit{Jilin University}\\
Jilin, China \\
amit2122@mails.jlu.edu.cn}
\and
\IEEEauthorblockN{2\textsuperscript{nd} Xiaoli Zhang}
\IEEEauthorblockA{\textit{College of Computer Science and Technology} \\
\textit{Jilin University}\\
Jilin, China \\
zhangxiaoli@jlu.edu.cn}
\and
\IEEEauthorblockN{3\textsuperscript{rd} Md Masum Billa Shagar}
\IEEEauthorblockA{\textit{College of Computer Science and Technology (College of Data Science)} \\
\textit{Taiyuan University of Technology}\\
Taiyuan, China \\
masum1201@gmail.com}
\and
\IEEEauthorblockN{4\textsuperscript{th} Zhaojun Liu*}
\IEEEauthorblockA{\textit{College of Computer Science and Technology} \\
\textit{Jilin University}\\
Jilin, China \\
zhaojun@jlu.edu.cn \\
*Corresponding author}
\and
\IEEEauthorblockN{5\textsuperscript{th} Xiongfei Li}
\IEEEauthorblockA{\textit{College of Computer Science and Technology} \\
\textit{Jilin University}\\
Jilin, China \\
lxf@jlu.edu.cn}
\and
\IEEEauthorblockN{6\textsuperscript{th} Fanlong Meng}
\IEEEauthorblockA{\textit{Department of Computer Science} \\
\textit{Delaware State University}\\
Dover, United States \\
felixmengg@gmail.com}
}

\maketitle

\begin{abstract}
    Effectively describing features for cross-modal 
    remote sensing image matching remains a 
    challenging task due to the significant 
    geometric and radiometric differences 
    between multimodal images.
    Existing methods primarily extract features 
    at the fully connected layer but often fail 
    to capture cross-modal similarities effectively.
    We propose a Cross Spatial Temporal Fusion 
    (CSTF) mechanism that enhances feature 
    representation by integrating scale-invariant 
    keypoints detected independently in both 
    reference and query images. Our approach 
    improves feature matching in two ways: First, 
    by creating correspondence maps that leverage 
    information from multiple image regions 
    simultaneously, and second, by reformulating 
    the similarity matching process as a 
    classification task using SoftMax and Fully 
    Convolutional Network (FCN) layers. This dual 
    approach enables CSTF to maintain sensitivity 
    to distinctive local features while 
    incorporating broader contextual information, 
    resulting in robust matching across diverse 
    remote sensing modalities. To demonstrate the 
    practical utility of improved feature matching, 
    we evaluate CSTF on object detection tasks 
    using the HRSC2016 and DOTA benchmark datasets. 
    Our method achieves state-of-the-art 
    performance with an average mAP of 90.99\% on 
    HRSC2016 and 90.86\% on DOTA, outperforming 
    existing models. The CSTF model maintains 
    computational efficiency with an inference 
    speed of 12.5 FPS. These results validate 
    that our approach to cross-modal feature 
    matching directly enhances downstream remote 
    sensing applications such as object detection.
\end{abstract}

\begin{IEEEkeywords}
    Cross Spatial Temporal Fusion, Attention Mechnasim, Transformer, Image Classification
\end{IEEEkeywords}

\section{Introduction}
Remote sensing image matching is a fundamental process 
that involves identifying and aligning two pixel-wise 
multi-temporal or multi-sensor images of the same scene. 
It plays a critical role in remote sensing image 
processing, enabling tasks such as image fusion, 
change detection, multi-temporal image analysis, 
and image stitching. Various satellite sensors 
provide remote sensing images across multiple 
spectral bands, captured at different times and 
resolutions (i.e., multimodal images), offering 
complementary information about the same geographic 
regions. To correct global geometric distortions—such as rotations 
and scale changes—between remote sensing images, 
current methods often rely on navigation 
parameters \cite{Ye2022ARM}. Image registration 
techniques can generally be classified into two 
main categories: feature-based methods and area-based 
methods \cite{Jiang2021ARO}. A well-known feature-based 
method is the Scale-Invariant Feature Transform 
(SIFT) \cite{Lowe2004DistinctiveIF}, which 
generates feature descriptors that remain 
consistent despite image translations, 
rotations, and scale changes. Variants 
like SAR-SIFT \cite{Dellinger2015SARSIFTAS}, 
DSP-SIFT, and RIFT \cite{Dong2014DomainsizePI} 
have been developed to improve the accuracy of 
local feature description.

However, these traditional methods often 
struggle with large-scale multimodal remote 
sensing image matching, where indistinct 
features are common. To overcome these limitations, 
many recent approaches have turned to deep learning 
for feature discovery and matching \cite{Zhou2022RobustMF}. 
Modern deep learning-based matching techniques leverage 
advanced algorithms to extract keypoints and feature 
descriptors, with notable examples including 
LIFT \cite{Yi2016LIFTLI}, DELF \cite{Noh2016LargeScaleIR}, 
D2-Net \cite{Dusmanu2019D2NetAT}, 
ASLFeat \cite{Luo2020ASLFeatLL}, and 
SuperGlue \cite{Sarlin2019SuperGlueLF}. For instance, 
MAP-Net \cite{Cui2021MAPNetSA} uses self-attention to 
extract SAR and optical image features, embedding 
high-level semantic information for cross-modal 
matching. However, despite its advancements, MAP-Net 
fails to capture the connection between keypoints 
and the global context, leading to missed feature 
descriptions.

Traditional area-based approaches rely on similarity 
metrics to match local features within a search window. 
Common similarity measures include 
SSD \cite{Hisham2015TemplateMU}, 
NCC \cite{Suri2010MutualInformationBasedRO}, MI, 
HOPC, CFOG, and SFcNet \cite{Zhang2019RegistrationOM}. 
MI \cite{Maes1997MultimodalityIR} and 
HOPC \cite{Ye2017RobustRO} are better equipped to 
handle radiometric variations compared to SSD and NCC, 
though template distortions limit their effectiveness 
in multimodal remote sensing registration. To address 
these issues, several deep neural networks—such as 
Goodness \cite{Hughes2020ADL}, twin U-Net with 
FFT \cite{Fang2021SAROpticalIM}, HardNet \cite{Brgmann2019MatchingOT}, 
and Siamese CNNs \cite{Merkle2017ExploitingDM}—have 
been introduced. These models aim to enhance local 
patch similarity measures through weight-sharing 
architectures like pseudo-Siamese or Siamese networks.
In contrast to pixel-by-pixel search methods, 
which are time-consuming and labor-intensive, 
recent advancements such as the semantic template 
matching framework \cite{Li2021ADL} have improved 
performance by fusing features to establish image 
correspondences. However, they often overlook 
cross-spatial feature similarity relationships, 
limiting their ability to effectively capture 
structural dependencies across multimodal images.

To address these challenges, we propose an encoder-decoder 
feature cross-fusion mechanism that integrates 
cross-spatial-temporal information. Our method enhances 
mutual information capture by aligning keypoints from both 
reference and sensed images while preserving spatial 
relationships across multiple scales. Specifically, 
the proposed cross-spatial-temporal module leverages 
spatial attention to enhance feature correspondence, 
ensuring robust alignment across multimodal images. 
This mechanism enables our model to extract 
distinctive keypoints while incorporating global 
contextual information.
Additionally, we introduce a novel similarity 
loss function that uses Fully Convolutional Network 
(FCN) layers and a dual SoftMax operation to measure 
matching confidence between cross-spatial-temporal 
feature descriptors. Unlike traditional similarity 
metrics, our approach formulates feature matching as 
a classification problem, allowing the model to learn 
more discriminative representations and improve 
robustness in multimodal image matching.

To demonstrate the practical utility of improved 
feature matching, we evaluate our Cross Spatial 
Temporal Fusion (CSTF) model on object detection 
tasks using the HRSC2016 and DOTA benchmark datasets. 
Our method achieves state-of-the-art performance, 
with an average mAP of 90.99\% on HRSC2016 and 
90.86\% on DOTA, outperforming existing models. 
The CSTF model maintains computational efficiency 
with an inference speed of 12.5 FPS. These results 
validate that improving cross-modal feature matching 
through our approach directly enhances downstream 
remote sensing applications such as disaster monitoring, 
urban planning, and environmental surveillance.

The contributions of this paper can be summarized 
as follows:

\begin{itemize}
\item We introduce a cross-spatial-temporal 
fusion mechanism that enhances multimodal image 
matching by integrating keypoint-based alignment 
with spatial attention, improving feature representation 
across different imaging conditions.
\item We develop a novel similarity loss function 
that utilizes FCN layers and a dual SoftMax operation, 
transforming similarity estimation into a classification 
task, which enhances discriminative feature learning.
\item We investigate the impact of patch sizes on 
feature matching accuracy, demonstrating that a 
balanced capture of local and global context 
significantly improves multimodal image matching.
\item Our CSTF model outperforms traditional 
methods such as R3Det, S2ANet, ReDet, Oriented 
RepPoints, Point RCNN, Oriented RCNN, CGD, OII, 
and LSKNet in recall and mAP across key datasets, 
showcasing its robustness in complex multimodal settings.
\end{itemize} 

\section{Related work}
In computer vision, feature matching—the process of identifying 
the same scenes or objects across multiple images—is both 
essential and challenging. Effective feature matching 
facilitates tasks such as image stitching \cite{nie2021unsupervised}, 
3D reconstruction \cite{ham2019computer}, and 
pose estimation \cite{engel2017direct, Forster2014SVOFS} 
in various remote sensing applications.

Over time, researchers have made significant contributions 
to improving feature matching techniques to find more 
reliable and frequent matches. Recent works have 
focused on improving feature extraction methods, 
with algorithms such as SURF (Speeded-Up Robust Features) 
remaining a common choice for feature extraction and image 
matching purposes. Luo et al. (2024) improved SURF 
with RANSAC algorithms for image matching in 
indoor environments, further advancing the state 
of traditional matching approaches \cite{Luo2024ResearchOI}. 
Additionally, methods like Stereo Matching based on 
attention and scale fusion have been proposed 
to enhance remote sensing image matching \cite{Wei2024StereoMM}.

Real-world scenarios often introduce complexity 
that can limit the performance of traditional 
matching schemes, resulting in numerous outliers. 
To handle these challenges, techniques like 
RANSAC \cite{Horn1981DeterminingOF} and state 
estimation \cite{Yan2019SelfSupervisedLO} are 
often employed to refine the results. In remote 
sensing, change detection and multi-scale analysis 
are vital, and recent studies have enhanced 
these methods through innovative approaches, 
such as multi-scale spatio-temporal perceptual 
attention networks \cite{Jia2024RemoteSC}, and 
semantic change detection using multitask 
networks \cite{Zuo2025MultitaskSN}.

The general process of local feature matching 
typically involves several steps: $(i)$ detecting 
key points of interest, $(ii)$ generating visual 
descriptors for these points, $(iii)$ matching 
descriptors using Nearest Neighbor (NN) 
search, $(iv)$ filtering out false matches, 
and $(v)$ estimating a geometric transformation. 
The classical feature matching pipeline, 
developed in the 2000s, often relied on 
SIFT \cite{Lowe2004DistinctiveIF}, with 
methods such as Lowe’s ratio test, mutual 
matching checks, and neighborhood consensus 
heuristics \cite{Tuytelaars2000WideBS, CechIeeeTO, Bian2017GMSGM, 5459459}. 
Transformation estimation was typically performed 
using robust solvers like RANSAC \cite{Fischler1981RandomSC, Raguram2008ACA}.

In recent years, deep learning approaches have 
focused on improving sparse feature detectors 
and local descriptors using data-driven methods 
based on CNNs \cite{DeTone2017SuperPointSI, Dusmanu2019D2NetAT, Ono2018LFNetLL, r2d2, Villani2008OptimalTO}. 
To enhance discriminative power, some studies 
have expanded the scope by incorporating regional 
features \cite{Luo2019ContextDescLD} or log-polar 
patches \cite{Ebel2019BeyondCR}. Other approaches 
aim to classify matches as inliers or 
outliers \cite{Yi2017LearningTF, Ranftl2018DeepFM, Zhang2019LearningTC, Brachmann2019NeuralGuidedRL}, 
focusing on match filtering rather than refining the 
matching process itself.

Our approach, in contrast, employs a learnable 
middle-end that simultaneously performs context 
gathering, matching, and filtering in an end-to-end 
framework. Traditional graph matching problems are 
often modeled as quadratic assignment problems, 
which are NP-hard and require expensive, complex 
solvers, making them impractical for many 
applications \cite{Loiola2007ASF}.

In our method, match pairs are treated as 
displacement vectors. An accurate match is 
characterized by a consistent displacement among 
neighboring matches. By using the nearest neighbor 
approach, we can identify the surrounding 
matches for the pair under analysis. Changes 
in scale and rotation in remote sensing imagery 
have minimal effect on the stability of 
neighboring point associations, as previous 
studies have shown \cite{Lin2018CODECB}. The 
displacement vector is normalized relative to 
image size, and the consistency of neighboring 
feature matches is evaluated through 
motion analysis.
\section{Methodology}
\begin{figure}[t]
    \centering
    \includegraphics[width=0.45\textwidth]{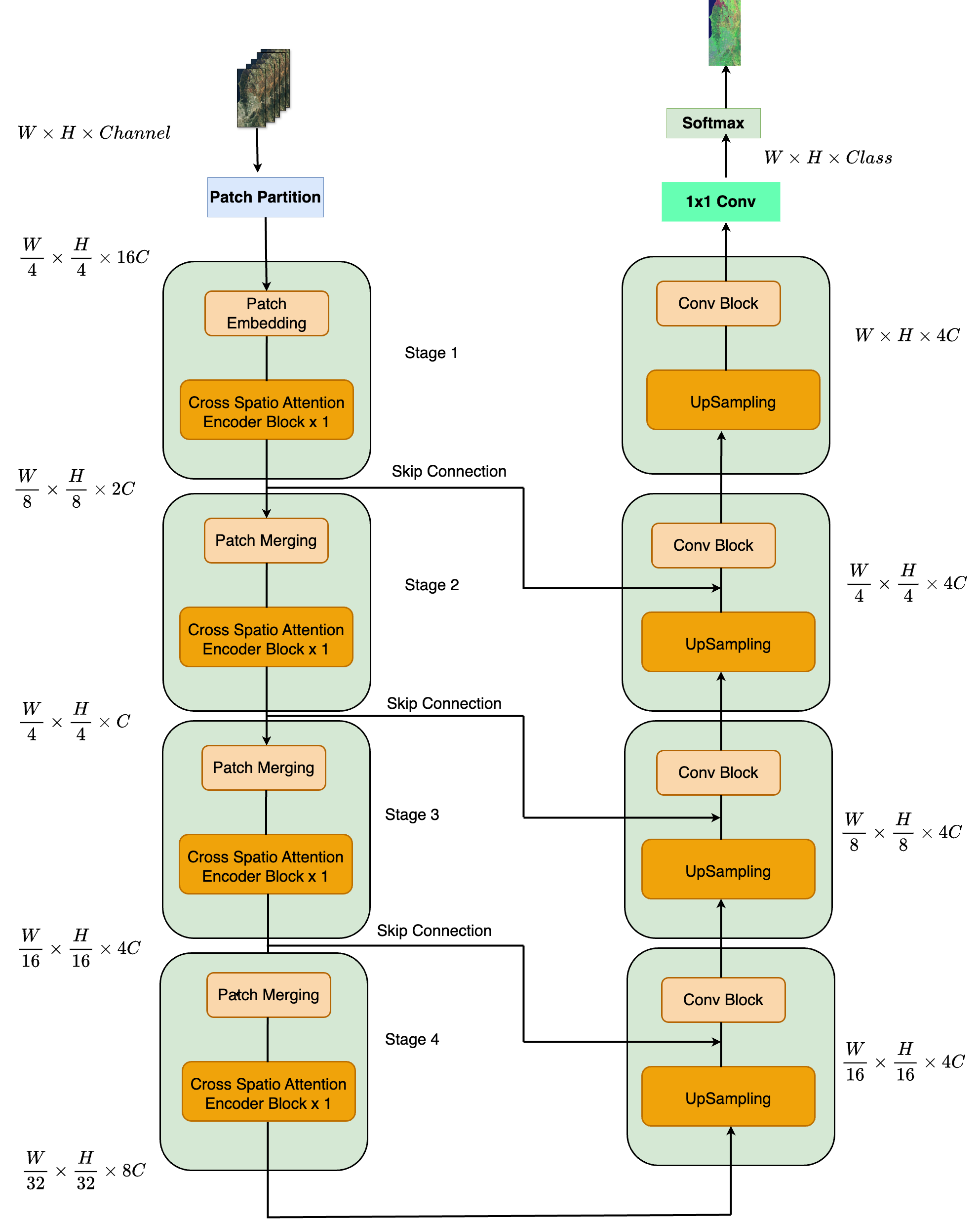}
    \caption{The proposed architecture of CSTF. First, image patches are embedded and split into $n$ patch tokens. These tokens are processed through the Cross Spatial Temporal Attention Fusion block to enhance deep temporal information interaction. The normalized outputs are then fed to the output layer.}\label{fig_1}
    \vspace*{-1.0\baselineskip}
\end{figure}

Figure \ref{fig_1} illustrates how our proposed Cross 
Spatial Temporal Fusion (CSTF) block is integrated into 
a fully convolutional network (FCN) with skip connections. 
The CSTF block maintains a flexible structure that remains 
consistent across different numbers of encoder stages. 
Given an FCN with $n+1$ multiscale encoder stages, the 
CSTF block processes features from the first $n$ stages, 
corresponding to the final convolutional outputs at each 
stage. It enhances these multiscale features before 
linking them to the $n$ corresponding decoder stages.

\begin{figure}[t]
    \centering
    \includegraphics[width=0.45\textwidth]{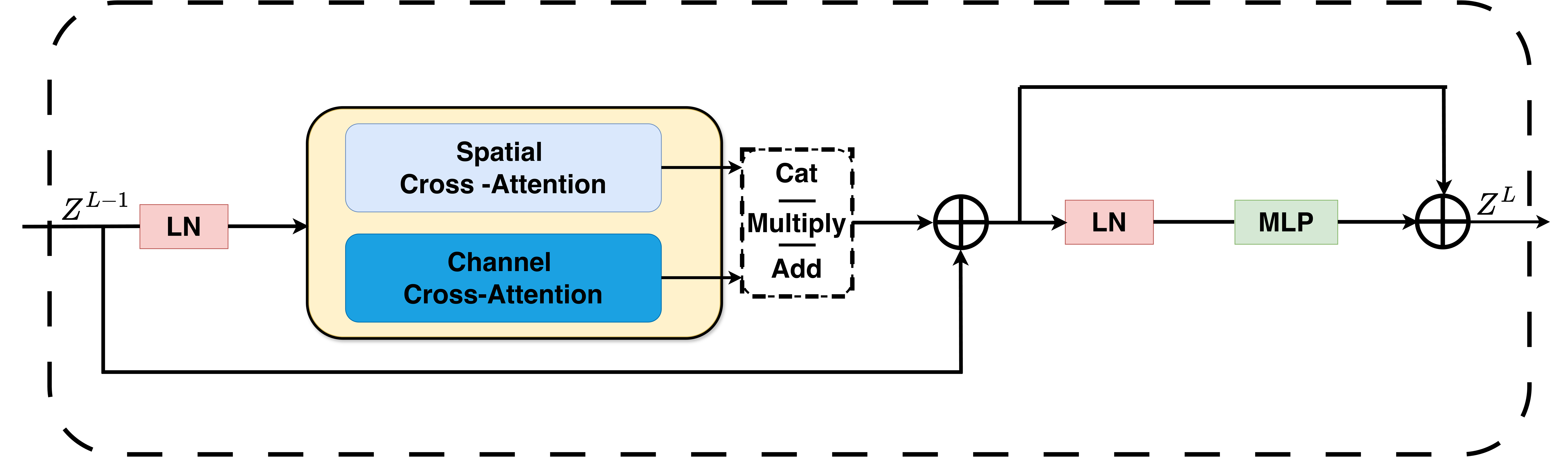}
    \caption{Cross Spatial Attention Encoder Block}\label{fig_2}
    \vspace*{-1.0\baselineskip}
\end{figure}

As shown in Figure \ref{fig_2}, the CSTF block 
operates in two distinct phases. In the first phase, a 
multiscale patch embedding module extracts encoder tokens 
from each stage. The second phase applies the CSTF mechanism, 
which integrates Cross Attention (CA) and Spatial Cross 
Attention (SCA) to process these tokens. This approach 
effectively captures long-range dependencies within 
the input data. After applying CSTF, the token sequence 
undergoes layer normalization and GeLU activation. 
Finally, the tokens are upsampled to higher resolutions, 
ensuring efficient feature propagation to the 
corresponding decoder stages for tasks such as segmentation.

\subsection{Patch Partition}\label{sec3partition}
Before extracting feature embeddings, the input feature maps 
are first divided into non-overlapping patches across multiple 
encoder stages. Given an encoder feature map at the $i$-th stage with resolution:

\begin{equation} 
    E_{i} \in \mathbb{R}^{C_{i} \times \frac{H}{2^{i-1}} \times \frac{W}{2^{i-1}}} 
\end{equation}

\noindent a partitioning operation divides it into patches 
of size $P_{i}^{s}$ using a progressive reduction factor:

\begin{equation} 
    P_{i}^{s} = \frac{P^{s}}{2^{\frac{i-1}{2}}} 
\end{equation}

where $i = 1,2, \cdots, n$. This adaptive 
scaling balances local detail retention with effective 
downsampling, ensuring that higher-resolution features 
maintain fine-grained spatial information crucial for 
accurate image matching.

Each partitioned region undergoes 2D average pooling 
with a pool size and stride of $P_{i}^{s}$, generating 
patch tokens that are flattened into 
vectorized representations:

\begin{equation} 
    T_{i} \in \mathbb{R}^{P \times C_{i}} 
\end{equation}

where $P$ is the total number of 
patches, kept consistent across all stages to 
maintain alignment in the fusion process.

\subsection{Patch Embedding}\label{sec2Embedding}

Once the patches are extracted, they are embedded 
into a latent feature space to capture hierarchical 
information and improve cross-scale matching. 
A depth-wise $1 \times 1$ convolution is applied for 
efficient feature projection:

\begin{equation}\label{eq4} 
    \hat{T}{i} = Conv{1 \times 1}^{i} (T_{i}) 
\end{equation}

where $\hat{T}_{i}$ denotes the transformed 
token representation for the $i$-th encoder stage. 
This step compresses spatial information while enhancing 
channel-wise interactions, preparing the tokens for 
Cross-Spatial Temporal Fusion (CSTF).

\subsection{Encoder with Multiscale Feature Embedding}

Our proposed Cross Spatial Temporal Fusion (CSTF) 
block is designed to enhance image matching and 
object detection by effectively integrating spatial, 
temporal, and cross-scale feature information within 
a fully convolutional network (FCN). The CSTF block 
refines multiscale features extracted from different 
encoder stages, ensuring robust feature representation 
for improved object localization and matching accuracy. 
The methodology consists of three main components: 
multiscale feature embedding, spatial-temporal fusion, 
and hierarchical decoding.

The encoder follows a hierarchical structure 
with $n+1$ multiscale stages, where the first $n$ encoder 
stages generate feature maps at different resolutions. 
Each encoder stage consists of convolutional layers, 
patch embedding, and cross-attention modules. Given a 
feature map at the $i$-th encoder stage:

\begin{equation}
E_{i} \in \mathbb{R}^{C_{i} \times \frac{H}{2^{i - 1}} \times \frac{W}{2^{i - 1}}}
\end{equation}

we apply adaptive patch extraction using a progressive scaling factor:

\begin{equation}
P_{i}^{s} = \frac{P^{s}}{2^{\frac{i-1}{2}}}
\end{equation}

Patches are extracted using 2D average pooling with a kernel size and stride of $P_{i}^{s}$, ensuring smooth transition across scales while preserving spatial details. Each patch is then flattened into a feature vector:

\begin{equation}
T_{i} \in \mathbb{R}^{P \times C_{i}}
\end{equation}

Projected into an embedding space via a $1 \times 1$ depth-wise convolution:

\begin{equation}
\hat{T}{i} = \text{Conv}{1 \times 1}^{i}(T_{i})
\end{equation}

These tokens are then processed by the CSTF mechanism 
for enhanced feature refinement.

\begin{figure}[t]
    \centering
    \includegraphics[width=0.40\textwidth]{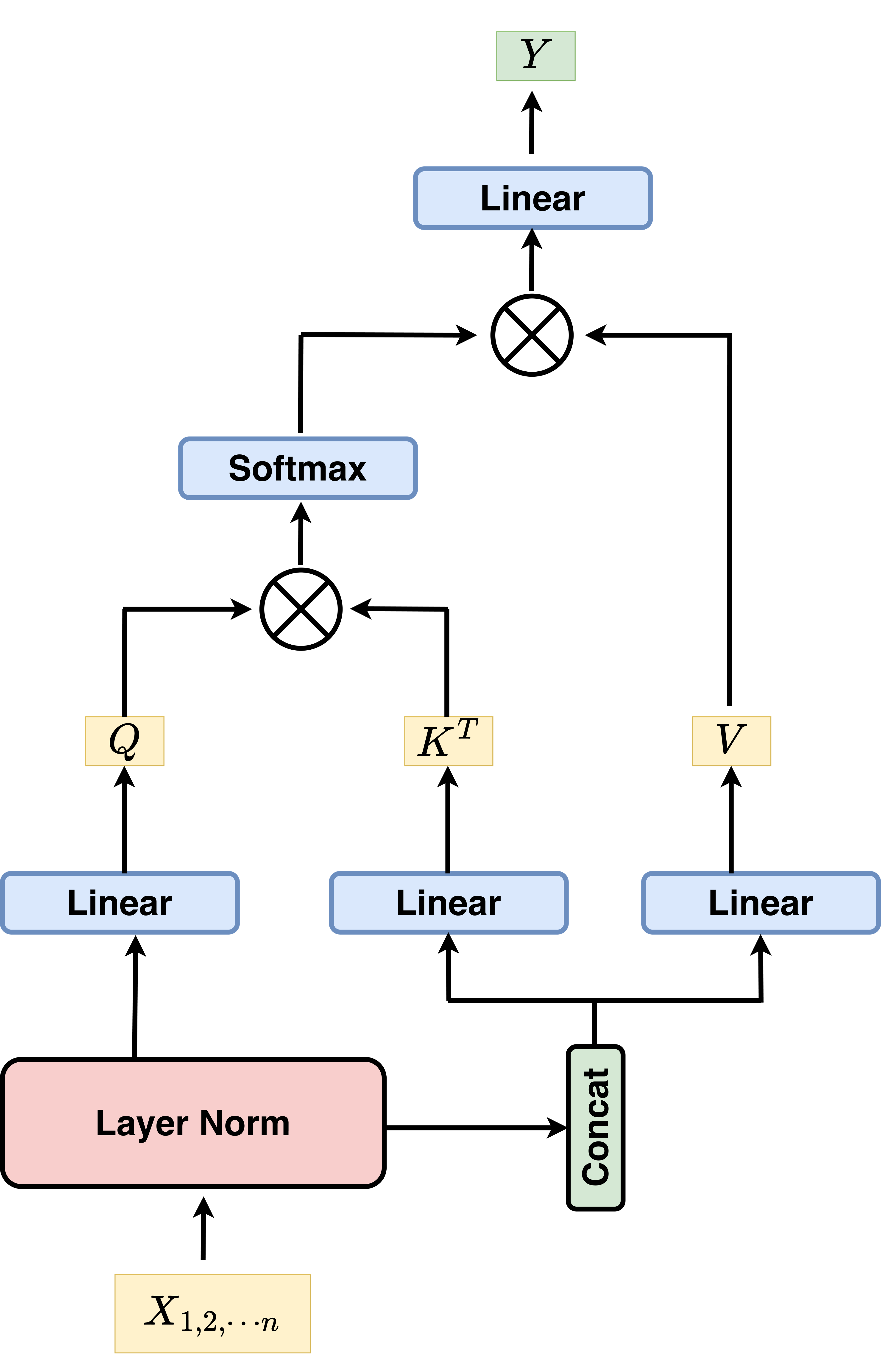}
    \caption{Channel Cross Attention}\label{fig_3}
    \vspace*{-1.0\baselineskip}
    \end{figure}
    \begin{figure}[t]
    \centering
    \includegraphics[width=0.40\textwidth]{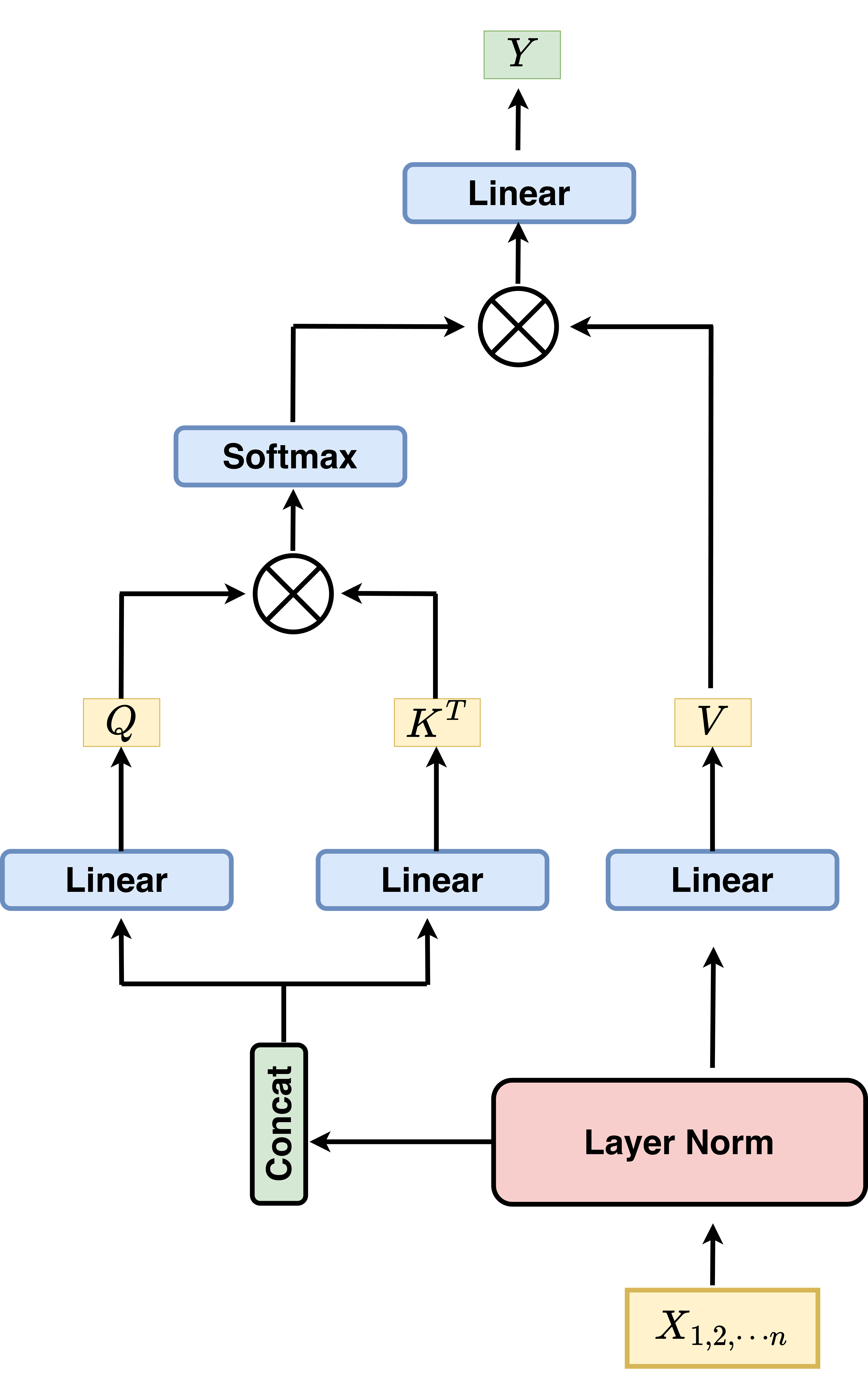}
    \caption{Spatial Cross Attention Module}\label{fig_4}
    \vspace*{-1.0\baselineskip}
\end{figure}

\subsection{Cross Spatial Temporal Fusion (CSTF)}
The CSTF module consists of two key operations:
Cross Attention (CA) – Captures dependencies across 
multiple encoder stages, reinforcing relationships 
between low-level and high-level features.
Spatial-Temporal Cross Attention (STCA) – Integrates 
intra-stage spatial information and temporal 
consistency, crucial for accurate image matching 
and object detection.

CA improves feature propagation between different encoder 
levels by computing attention weights from query, key, and 
value projections:

\begin{equation}
Q_{i} = W_{Q} \hat{T}{i}, \quad K_i = W{K} \hat{T}{i}, \quad V{i} = W_{V} \hat{T}_{i}
\end{equation}

\begin{equation}
\text{Attention}{i} = \text{Softmax}\left(\frac{Q_i K{i}^{T}}{\sqrt{\tilde{C}_i}}\right)V_i
\end{equation}

Resulting enriched tokens:

\begin{equation}
\hat{T}'{i} = \sum{i} \text{Attention}_{i}
\end{equation}

The Cross Spatial Temporal Fusion procedure begins 
with extracting multiscale patches from multiple 
encoder stages, each operating at different spatial 
resolutions. For each encoder stage, feature maps are 
divided into patches through a 2D average pooling 
process, scaled according to the resolution of the 
encoder stage. These patches are flattened into 
vectors and projected into an embedding space 
using a $1 \times 1$ depth-wise convolution, 
creating tokens representing the features at each scale.

Next, CA is applied to capture dependencies 
across channels within and between the encoder 
stages. Each token from the encoder stages is 
projected into query, key, and value representations,
and cross-attention is computed by comparing similarities 
between queries and keys. The resulting attention weights 
update the values, producing enhanced tokens containing 
cross-channel information from multiple stages.

Spatio-Temporal Cross Attention (STCA) refines 
feature consistency by incorporating 
spatial and temporal dependencies within each encoder stage. 

Following the CA module, Self-Cross Attention (SCA) 
refines tokens by focusing on intra-stage dependencies. 
Similar to CA, tokens are projected into new query, key, 
and value representations, and self-attention is applied 
to capture important intra-stage relationships. Tokens 
are up-projected back to their original dimensions after 
the self-attention step.

Finally, the output tokens from the SCA module are 
combined with the original tokens from the encoder 
stages using residual connections, retaining the 
original multiscale features while incorporating 
enhanced representations generated through CA and 
SCA. These fused tokens are then passed to the decoder 
stages for tasks like segmentation, leveraging 
cross-scale and cross-channel information for improved 
spatial-temporal modeling.

The encoder incorporates two consecutive transformer blocks: 
the channel cross-attention module and the spatial 
cross-attention module. Figures \ref{fig_3} and 
\ref{fig_4} illustrate these blocks.
The input of the current block is the output 
feature $Z^{L - 1}$ of the previous encoder block. 
Linear normalization (LN) and the input are applied 
to the cross-spatial attention encoder by $Z^{L - 1}$. 
Spatial cross-attention is computed as described earlier, 
with output obtained through a fusion mechanism. Channel 
cross-attention operates as follows:

\begin{equation}
Q_{i} = W_{Q} \hat{T}{i}, \quad K_i = W{K} \hat{T}{i}, \quad V{i} = W_{V} \hat{T}_{i}
\end{equation}

\begin{equation}
\text{Attention}{i} = \text{Softmax}\left(\frac{Q_i K{i}^{T}}{\sqrt{\tilde{C}_i}}\right)V_i
\end{equation}

Where $C_{i}$ is the number of channels in the $i$-th stage. Enhanced tokens enriched with cross-channel information are computed as:

\begin{equation}
\hat{T}'{i} = \sum{i} \text{Attention}_{i}
\end{equation}

Following CA, the tokens undergo further 
refinement through the SCA module. 
This module focuses on intra-stage dependencies:

\begin{equation}
Q'i = W{Q'} \hat{T}_i', \quad K'i = W{K'} \hat{T}_i', \quad V'i = W{V'} \hat{T}_i'
\end{equation}

\begin{equation}
\text{SCA}_{i} = \text{Softmax}\left(\frac{Q'_i (K'_i)^T}{\sqrt{\tilde{C}_i}}\right)V'_i
\end{equation}

\begin{equation}
\tilde{\hat{T}}''{i} = W{up}^{i} \text{SCA}_{i}
\end{equation}

Tokens are fused back into the original feature maps 
using residual connections:

\begin{equation}
\hat{T}''{i} = \hat{T}'{i} + T_{i}
\end{equation}

Final tokens $\hat{T}''_{i}$ are passed to the 
decoder stages for tasks such as segmentation, 
leveraging comprehensive multiscale feature fusion.

\subsection{Hierarchical Decoder with Feature Aggregation}

The decoder reconstrcuts high\-resolution feature
representations by progressively upsampling 
the encoded features while 
integrating information form skip connections. 
It follows a hierarchical structure 
that mirrors the encoder, enabling effective feature fusion and 
spatial detail preservation.
Each decoder stage consist of upsampling which expands the 
spatial resolution of feature maps. Convolution blocks which refines feature represenation and skip connections fuses features from 
the encoder to enhance detail retention.

Formally, for the $i$-th decoder stage,
the upsampling operation restores the spatial resolution: 

\begin{equation}
    D_{i} = \text{UpSample}(D_{i+1}) \in \mathbb{R}^{C_{i} \times \frac{H}{2^{i-1}} \times \frac{W}{2^{i-1}}}
\end{equation}

where $D_{i+1}$ is the feature representation from the previous decoder 
stage. A Convolution transformation further refines these upsampled features:

\begin{equation}
    \hat{D}_{i} = \text{ConvBlock}(D_{i})
\end{equation}

Each decoder stage integrates corresponding 
encoder features via skip connections:

\begin{equation}
    \tilde{D}_{i} = \hat{D}_{i} + Z_{i}
\end{equation}

where $Z_{i}$ is the output from the corresponding encoder 
stage, ensuring that 
spatial details from the earlier stages are preserved.

Finally, the restored high-resolution feature map is passed through a $1 \times 1$
convolution to generate the final prediction:

\begin{equation}
    O = \text{Conv}_{1 \times 1}(\tilde{D}_{1})
\end{equation}

A softmax layer is applied to obtain 
the final output, ensuring a normalized probability distribution in segmentation or 
classification tasks.

The hierarchical upsampling approach, combined with convolutional 
refinement and conder-decoder fusion, enchances the model's ability 
to recover spatial details and maintain semantic consistency.

\section{Experiments}\label{sec4exp}

In this section, we evaluated our method on the  benchmark datasets 
and compared it to other widely used benchmarks 
such as IFDet \cite{He2024IFDetIA}, S2ANet \cite{Han2020AlignDF}, 
ReDet \cite{han2021ReDet}, Oriented RCNN \cite{Xie2021OrientedRF}, 
and LSKNet \cite{Li2023LargeSK}.
Subsequently, we provide the experimental datasets, 
which encompass various categories of remote-sensing photos. Lastly, 
we meticulously explain the qualitative, quantitative, and resilient 
experimental outcomes obtained through our methodology.

\subsection{Datasets}
The HRSC2016 \cite{Liu2016ShipRB} dataset serves as a standard benchmark 
for remote sensing ship detection, featuring a 
range of intricate scenes such as coastal, offshore, 
and port areas. The ships in this dataset vary in size, 
orientation, and shape, while the background encompasses 
diverse elements like oceans, ports, and islands. On the 
other hand, the DOTA \cite{Xia2017DOTAAL} dataset is among the most commonly 
used benchmarks for remote sensing object detection tasks. 
It includes high-resolution images taken from different altitudes 
and perspectives, covering complex environments such as urban 
landscapes, forests, oceans, and busy ports. We utilize this 
dataset as a validation set to assess the 
generalization performance of our model.

\subsection{Evaluation Metrics}
We evaluate the performance of our method 
using Mean Average Precision (mAP) and 
recall, two standard metrics in object 
detection tasks derived from the 
PASCAL VOC evaluation standard.
mAP is a widely used metric in object detection that 
measures the accuracy of the model in detecting objects 
across different classes. It is calculated as the 
average of the Average Precision (AP) for each class. 
AP is computed by first generating a precision-recall 
curve for each class, which plots precision 
(the percentage of correct detections among 
all detections) against recall (the 
percentage of correct detections among 
all ground truth instances) at various 
confidence thresholds. The area under this 
curve gives the AP, and the mAP is the mean 
of AP values across all classes..
mAP07 refers to the mean Average 
Precision at the 0.7 IoU (Intersection over Union) 
threshold, which is a stricter evaluation criterion. 
It requires that the predicted bounding box and 
the ground truth bounding box overlap by at 
least 70\% for a detection to 
be considered correct.
mAP12 refers to the mean Average 
Precision at the 0.5 IoU threshold, 
which is a more lenient evaluation 
criterion, allowing for a smaller 
overlap between the predicted and ground truth boxes.
Recall, also known as True Positive Rate (TPR), 
is a measure of the model's ability to 
detect all relevant objects in the 
dataset. It is calculated as the 
ratio of true positive detections to 
the total number of ground truth objects. 
A high recall indicates that the model 
is good at identifying objects, but it 
may also result in false positives.
Additionally, the evaluation of the 
model's performance is conducted by 
comparing it across various parameters, 
including Params, floating-point operations 
(Flops), and frames per second (FPS), to 
assess its computational efficiency 
and runtime performance.

\begin{figure*}[t]
    \centering
    \includegraphics[width=\textwidth,center]{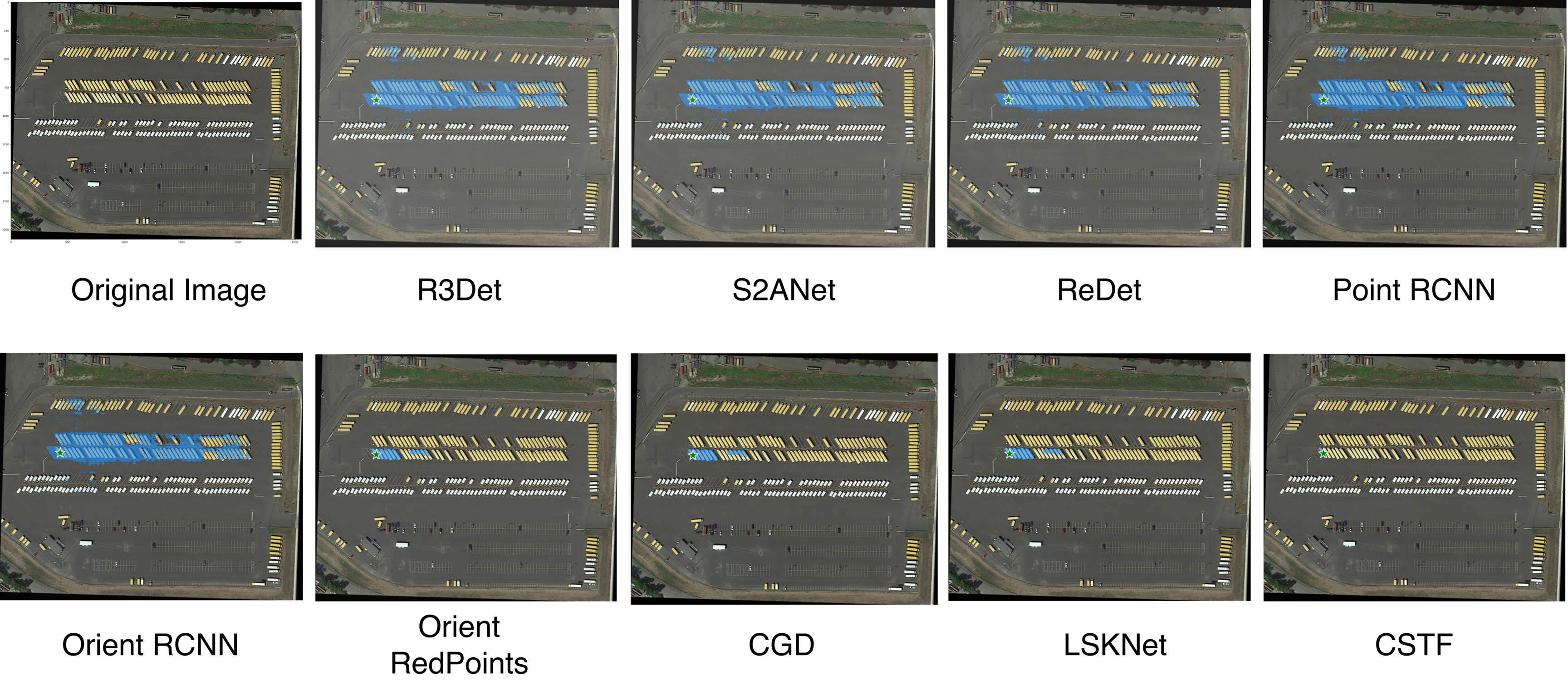}
    \caption{Visual representation of the model’s performance}
    \label{fig_visual_img}
\end{figure*}

\subsection{Comparison with state of the art methods}

To assess the effectiveness of our proposed Cross Spatial 
Temporal Fusion (CSTF) approach, we conducted extensive 
comparative experiments on the HRSC2016 and DOTA 
datasets against state-of-the-art object detection 
models, including IFDet, S2ANet, ReDet, Oriented RCNN, and 
LSKNet. The results, summarized in Tables \ref{tab3_new_HRSC2016} 
and \ref{tab4_new_Dota}, demonstrate the superior 
performance of our method in terms of both detection 
accuracy and recall.

On the HRSC2016 dataset, our CSTF method achieves the 
highest performance across all metrics. CSTF attains 98.50\% 
recall, outperforming LSKNet (97.87\%) and Oriented 
RCNN (97.50\%). For mAP (VOC2007), CSTF reaches 90.70\%, 
exceeding LSKNet (90.61\%) and Oriented RCNN (90.53\%). 
In terms of mAP (VOC2012), CSTF achieves 98.46\%, demonstrating 
superior detection accuracy compared to ReDet (96.63\%) and 
IFDet (96.01\%). These results confirm that our model 
effectively enhances feature fusion and spatial-temporal 
dependencies, leading to improved detection precision in 
complex remote sensing scenarios.

We further evaluate our model's generalization ability 
on the DOTA dataset, where CSTF again achieves 
state-of-the-art results. CSTF attains 96.30\% recall, 
significantly surpassing LSKNet (89.87\%) and Oriented 
RCNN (86.00\%). For mAP (VOC2007), CSTF scores 95.70\%, 
outperforming LSKNet (94.61\%) and Oriented RCNN (91.53\%). 
CSTF achieves 95.66\% for mAP (VOC2012), further 
highlighting its robustness in large-scale aerial 
image analysis.

\begin{table}
    \centering
    \caption{Comparison of our method with other state-of-the-art methods on the HRSC2016 dataset.}
        \begin{tabular}{ccccc} 
        \hline
        Method & Recall07 & mAP07 & mAP12 \\ 
        \hline
        IFDet & 91.50 & 89.12 & 96.01 \\
        S2ANet & 93.50 & 89.95 & 95.01 \\
        ReDet & 95.50 & 90.17 & 96.63 \\
        Orient RCNN & 97.50 & 90.53 & 96.60 \\        
        LSKNet & 97.87 & 90.61 & 97.46 \\
        CSTF & 98.50 & 90.70 & 98.46 \\ 
        \hline
        \end{tabular}    
    \label{tab3_new_HRSC2016}
    \vspace*{-1.0\baselineskip}
\end{table}

\begin{table}
    \centering
    \caption{Comparison of our method with other state-of-the-art methods on the DOTA dataset.}
        \begin{tabular}{ccccc} 
        \hline
        Method & Recall07 & mAP07 & mAP12 \\ 
        \hline
        IFDet & 84.50 & 89.12 & 87.01 \\
        S2ANet & 85.23 & 89.95 & 88.01 \\
        ReDet & 85.50 & 90.17 & 89.63 \\
        Orient RCNN & 86.00 & 91.53 & 91.60 \\        
        LSKNet & 89.87 & 94.61 & 94.46 \\
        CSTF & 96.30 & 95.70 & 95.66 \\ 
        \hline
        \end{tabular}   
    \label{tab4_new_Dota}
    \vspace*{-1.0\baselineskip}
\end{table}

In addition to detection accuracy, we evaluate our 
model's computational efficiency by comparing parameter 
size, FLOPs, and inference speed (FPS) with other leading 
detection models (Table \ref{tab5_new}). CSTF has 42.98M 
parameters, slightly higher than IFDet (41.87M) and 
Oriented RCNN (41.13M), but optimized for performance. 
CSTF achieves the lowest FLOPs (168.09G), outperforming 
all compared methods, including IFDet (335.04G) and 
S2ANet (196.21G). This efficiency is attributed to our 
Adaptive Rotated Dynamic Convolution module, which 
reduces redundant computations while enhancing feature 
learning. Regarding inference speed, CSTF runs at 12.5 FPS, 
demonstrating competitive real-time performance, 
second only to S2ANet (15.3 FPS) while significantly 
outperforming IFDet (8.3 FPS).

\begin{table*}[t]
    \centering
    \caption{A comparison of the Params, FLOPs, and FPS of different rotated object detection models.}
    \begin{tabular}{cccccc}   
        \hline      
        
        & S2ANet & IFDet & LSKNet & Oriented RCNN & CSTF \\ 
        \hline
        Params(M) & 38.5 & 41.87 & 36.60 & 41.13 & 42.98 \\ 
        FLOPs(G) & 196.21 & 335.04 & 194.24 & 198.53 & 168.09 \\ 
        FPS & 15.3 & 8.3 & 11.4 & 11.9 & 12.5  \\
        \hline    
    \end{tabular}    
    \label{tab5_new}
\end{table*}

To further validate the robustness of our model, we perform 
a qualitative visual analysis using challenging scenes from 
the DOTA dataset (Figure \ref{fig_visual_img}). In dense 
small-target environments, competing methods like IFDet and 
ReDet exhibit false negatives due to their limited ability 
to differentiate overlapping objects. In cluttered backgrounds, 
models such as Oriented RCNN fail to recognize objects near 
edges due to inadequate spatial context modeling. CSTF 
significantly improves detection accuracy in these challenging 
conditions by integrating a coordinate-aware pyramid feature 
aggregation module into a CNN-Transformer hybrid 
architecture, enhancing fine-grained object representation.

Our results highlight that CSTF achieves the best balance 
between accuracy, efficiency, and real-time performance, 
making it a compelling choice for remote sensing object 
detection. By effectively capturing spatial-temporal 
dependencies and optimizing feature extraction, CSTF 
outperforms state-of-the-art models while maintaining 
lower computational costs.

\subsection{Effect of patch size}

In this section, we explore the impact 
of patch size 
on object detection performance. In object detection 
tasks, using a single pixel as the keypoint for detecting 
objects is insufficient, as it lacks the 
contextual and semantic 
information necessary for distinguishing 
features. A single pixel provide insufficient detail 
to capture the spatial relationships between objects or 
their surrounding environment. Thus, expanding the 
patch size to incorporate surrounding pixels is 
essential for improving the model's ability to 
identify and localize objects effectively.

To evaluate the effect of patch 
size on object detection performance, we 
tested sizes 
of $9 \times 9, 13 \times 13, 17 \times 17$, 
and $21 \times 21$ 
across the HRSC2016 and DOTA datasets. 
We maintained consistent training 
and testing hyperparameters across all 
experiments to ensure a fair comparison. The primary 
evaluation metric used was the Mean Average Precision (mAP),
which measures the model's overall performance in detecting objects. 
As shown in Table \ref{tab_4_patch}, detection accuracy 
generally improved with larger patch sizes, with 
the best mAP observed at $21 \times 21$ for 
both datasets.

\begin{table}[t]
    \centering
    \caption{ Patch size performance}  
    \begin{tabular}{ccc}   
    \hline 
    Patch Size & HRSC2016(mAP) & DOTA(mAP)  \\
    \hline
    $9 \times 9$ & 88.12 & 87.64 \\

    $13 \times 13$ & 89.38 & 89.62 \\    

    $17 \times 17$ & 90.12 & 90.21 \\

    $21 \times 21$ & \textbf{90.99} & \textbf{90.86} \\
    \hline 
    \end{tabular}    
    \label{tab_4_patch}
    \vspace*{-1.0\baselineskip}
\end{table}

However, the optimal patch 
size of $21 \times 21$ 
depends on the characteristics of each 
dataset, particularly the object scale and density.
The patch embedding 
mechanism, as described in Section \ref{sec2Embedding}, 
applies a progressive reduction factor, detailed in 
eq. \ref{eq4}, which facilitates smooth transitions 
across different scales, retaining more local details 
in higher-resolution stages while enabling effective 
downsampling.

While larger patches provide more contextual 
information, their effectiveness is influnced by factors such as 
image resolution and object distribution. For the HRSC2016 
dataset, which involves ship detection in coastal and 
port environments, larger patches help to capture the broader 
spatial patterns associated with ships, leading to better 
localization and classification. In contrast, the DOTA 
dataset, which contains objects in urban, forested, and 
industrial areas, may benefit from smaller patches, 
particularly for objects with fine details or those 
located in dense scenes. In such cases, smaller patches 
may be more effective in maintaining the structural 
integrity of smaller or more densely packed objects.

For example, the HRSC2016 dataset, 
which focuses on detecting ships, showed consistent 
improvements in mAP with patch sizes between $13 \times 13$ 
and $21 \times 21$.
In contrast, the DOTA dataset, with its diverse 
set of object categories, required more careful 
patch size selection. In complex multimodal 
scenarios, such as SAR-optical 
image matching, overly large patches may capture 
excessive background information, introducing noise into 
detection, while smaller 
patches might fail to provide sufficient context 
for accurate object detection.

\subsection{Ablation Study}
To analyze the effectiveness of various components within 
the Cross Spatial Temporal Fusion (CSTF) model, we conducted 
an ablation study on two benchmark datasets: HRSC2016 and 
DOTA. This study evaluates the impact of Cross Attention 
(CA), Spatial Cross Attention (SCA), and different fusion 
strategies on object detection performance. The results 
in Table \ref{tab_4} provide a comprehensive comparison 
of different CSTF configurations.
We first examine the individual contributions of CA and 
SCA by evaluating models where each component operates 
independently. The results indicate that CSTF-(CA) 
achieves 89.32\% mAP on HRSC2016 and 89.64\% on DOTA. 
While CA captures fine-grained feature correspondences 
between keypoints, its reliance on pairwise relationships 
limits its ability to model global spatial dependencies 
effectively. CSTF-(SCA) performs slightly better, with 89.86\% 
mAP on HRSC2016 and 90.01\% on DOTA. This improvement 
suggests that spatial dependency modeling enhances object 
localization in complex scenes. Despite their effectiveness, 
neither CA nor SCA alone fully captures both fine-grained 
and global features, leading to suboptimal performance 
in object detection.

To improve feature interaction, we explore three 
different fusion strategies. In the Summation Fusion 
approach (CSTF-CA+SCA), CA and SCA outputs are computed 
in parallel and summed. While this approach enhances 
multi-level feature interaction, it limits the model's 
flexibility in learning fine-grained spatial relationships, 
yielding 91.33\% and 91.02\% mAP on HRSC2016 and DOTA, 
respectively. For the Concatenation Fusion strategy 
(CSTF-CA$||$SCA), features from CA and SCA are concatenated 
before further processing. This method achieves the best 
overall performance, with 92.42\% mAP on HRSC2016 and 92.16\% 
on DOTA. The improved results highlight the advantage of 
retaining distinct attention features, which allows richer 
spatial and semantic information. In the Sequential Fusion 
approach (CSTF-CA-SCA), CA is applied first, followed by 
SCA in a sequential manner. This strategy underperforms, 
as the sequential dependency hinders simultaneous feature 
learning, resulting in a performance drop. The superior 
performance of CSTF-CA$||$SCA suggests that retaining 
distinct feature representations before fusion enhances 
feature discrimination and robustness.

We also compare Convolutional Patch Embedding (CSTF-Conv) 
against 2D Average Pooling (CSTF-AP) for feature extraction. 
CSTF-Conv achieves 89.31\% mAP on HRSC2016 and 89.25\% on DOTA. 
While convolutional embedding extracts fine-grained local 
features, it struggles with long-range dependencies. 
CSTF-AP shows slightly better performance, with 89.42\% mAP 
on HRSC2016 and 89.31\% on DOTA. The global averaging helps 
retain broader spatial information, making it particularly 
effective for datasets like DOTA, which contain diverse 
object scales and orientations.

From this ablation study, we can draw several key takeaways. 
Both CA and SCA contribute to feature learning, but their 
combination significantly enhances object detection 
performance. Concatenation fusion (CSTF-CA$||$SCA) is the 
most effective strategy, providing the best balance of 
fine-grained and global spatial features. Summation 
fusion (CSTF-CA+SCA) slightly reduces performance due 
to the restrictive nature of additive feature blending. 
Sequential fusion (CSTF-CA-SCA) introduces bottlenecks, 
making it less effective for complex object structures. 
Additionally, 2D Average Pooling (CSTF-AP) outperforms 
Convolutional Embedding (CSTF-Conv) due to better spatial 
preservation. These findings confirm that the proposed 
CSTF-CA$||$SCA fusion strategy optimally leverages both 
cross-attention and spatial dependencies, significantly 
improving feature matching and object detection performance.

\begin{table}[t]
    \centering  
    \caption{ Quantitative comparison of CSTF Network ablation}
    \begin{tabular}{ccc}   
    \hline 
    Network & HRSC2016(mAP) & DOTA(mAP)  \\
    \hline
    CSTF-(CA) & 89.32 & 89.64 \\

    CSTF-(SCA) & 89.86 & 90.01   \\    

    CSTF-CA+SCA & 91.33  & 91.02 \\

    CSTF-CA$||$SCA & 92.42  & 92.16  \\

    CSTF-Conv & 89.31 & 89.25\\

    CSTF-AP & 89.42 & 89.31 \\
    \hline 
    \end{tabular}    
    
    \label{tab_4}
    \vspace*{-1.0\baselineskip}
\end{table}

\subsection{Limitations}
The CSTF model performs well in object detection 
tasks across datasets like HRSC2016 and DOTA, 
but several limitations require attention. 
One key issue is its dependence on dataset variability. 
While it works well on datasets with controlled spatial 
and temporal variations, its performance in extreme 
conditions—such as dense cloud cover or distorted 
terrains—remains uncertain. The model also faces 
challenges in highly multimodal contexts, where 
significant spectral fluctuations occur, as seen 
in the DOTA dataset with its diverse modalities 
(optical, infrared, radar). This suggests the need 
for improved strategies to handle spectral 
variations effectively.

Another constraint is the model’s computational 
complexity. Contextual attention and patch-based 
processing introduce considerable overhead, 
making it less suitable for real-time 
applications or environments with limited 
computational resources. Additionally, the model’s 
performance is sensitive to patch size and 
hyperparameters, which can be challenging 
to optimize in real-world scenarios where 
extensive tuning isn’t feasible. This is 
particularly true for datasets like HRSC2016 
and DOTA, where object scales and scene complexity 
vary, requiring careful configuration 
for optimal results.

\section{Conclusion}

We propose the Cross-Spatial Temporal Fusion (CSTF) 
model, incorporating an attention mechanism to 
enhance object detection accuracy in remote 
sensing. By transforming the detection task into a 
classification problem using SoftMax and Fully 
Connected Network (FCN) layers, CSTF refines 
feature representation, reduces noise, and enhances 
robustness in multimodal and complex environments.
Extensive experiments on the HRSC2016 and DOTA 
datasets demonstrate CSTF's competitive 
performance. On HRSC2016, it achieves an mAP of 0.814, 
while on DOTA, it reaches a mAP of 0.335. These 
results suggest that CSTF outperforms several 
existing object detection methods, including YOLO, 
Faster R-CNN, and SSD, particularly in handling 
complex spatial patterns and diverse object scales.

However, while CSTF shows promise in improving 
accuracy and efficiency for ship detection (HRSC2016) 
and object detection in urban and rural 
environments (DOTA), its performance in extreme 
conditions—such as dense cloud cover or 
distorted terrains—and scalability to 
larger scenes or long-term temporal 
analysis still require further investigation. 
The model lays a foundation for advanced 
object detection but highlights the need 
for future work to address its limitations.

\bibliographystyle{plain}
\bibliography{refs}
\vspace{12pt}
\color{red}

\end{document}